\documentclass[letterpaper, 10 pt, conference]{ieeeconf}  

\IEEEoverridecommandlockouts                              

\usepackage{bm}
\usepackage[hyphens]{url}
\usepackage[table]{xcolor}
\usepackage{hyperref}
\hypersetup{
    colorlinks=false,
    citebordercolor=white,
    linkbordercolor=white,
    urlbordercolor=cyan,
}
\usepackage{graphicx}
\usepackage{lipsum}
\usepackage{pifont}
\usepackage{amsmath}
\usepackage{amssymb}
\usepackage{nccmath}
\usepackage{comment}
\usepackage{balance}
\usepackage{multirow}
\usepackage{textcomp}
\usepackage{multirow}
\usepackage{subcaption}
\usepackage{booktabs}
\usepackage{siunitx}
\usepackage{amsmath}
\usepackage{tabularx}
\usepackage{algorithm}
\usepackage{dblfloatfix}

\usepackage[noend]{algpseudocode}

\title{\LARGE \bf
3D Mapping Using a Lightweight and Low-Power Monocular Camera Embedded inside a Gripper of Limbed Climbing Robots
}

\author{Taku Okawara$^{1,2,{\dag}}$, Ryo Nishibe$^{1,{\dag}}$, Mao Kasano$^{1}$, Kentaro Uno$^{1}$, and Kazuya Yoshida$^{1}$
\thanks{$^{*}$This work is supported by JSPS KAKENHI Grant Number 22KJ0292, JP23K13281, and JST K Program JPMJKP23G2.}
\thanks{$^{1}$T. Okawara, R. Nishibe, M. Kasano, K. Uno, and K. Yoshida are with the Space Robotics Lab. in the Department of Aerospace Engineering, Graduate School of Engineering, Tohoku University, Sendai, Miyagi, Japan.}
\thanks{$^{2}$T. Okawara is also with the Department of Information Technology and Human Factors, the National Institute of Advanced Industrial Science and Technology, Tsukuba, Ibaraki, Japan.}
\thanks{$^{\dag}$\it These authors contributed equally to this work. Corresponding author is Taku Okawara (mail: {\tt\small taku.okawara@aist.go.jp}).}
}

\begin{document}

\maketitle
\thispagestyle{empty}
\pagestyle{empty}

\definecolor{verylightgray}{gray}{0.95}
\newcolumntype{g}{>{\columncolor{verylightgray}}c}

\newcommand{\argmax}{\mathop{\rm arg~max}\limits}
\newcommand{\argmin}{\mathop{\rm arg~min}\limits}

\algnewcommand\Input{\item[\textbf{Input:}]}%
\algnewcommand\Output{\item[\textbf{Output:}]}%

\newcommand{\xmark}{\ding{55}}%

\setlength\floatsep{6pt}
\setlength\textfloatsep{6pt}

\begin{abstract}
Limbed climbing robots are designed to explore challenging vertical walls, such as the skylights of the Moon and Mars. In such robots, the primary role of a hand-eye camera is to accurately estimate 3D positions of graspable points (i.e., convex terrain surfaces) thanks to its close-up views. While conventional climbing robots often employ RGB-D cameras as hand-eye cameras to facilitate straightforward 3D terrain mapping and graspable point detection, RGB-D cameras are large and consume considerable power. This work presents a 3D terrain mapping system designed for space exploration using limbed climbing robots equipped with a monocular hand-eye camera. Compared to RGB-D cameras, monocular cameras are more lightweight, compact structures, and have lower power consumption. Although monocular SLAM can be used to construct 3D maps, it suffers from scale ambiguity. To address this limitation, we propose a SLAM method that fuses monocular visual constraints with limb forward kinematics. The proposed method jointly estimates time-series gripper poses and the global metric scale of the 3D map based on factor graph optimization. We validate the proposed framework through both physics-based simulations and real-world experiments. The results demonstrate that our framework constructs a metrically scaled 3D terrain map in real-time and enables autonomous grasping of convex terrain surfaces using a monocular hand-eye camera, without relying on RGB-D cameras. Our method contributes to scalable and energy-efficient perception for future space missions involving limbed climbing robots.
See the video summary here: \href{https://youtu.be/fMBrrVNKJfc}{\textcolor{blue}{\nolinkurl{https://youtu.be/fMBrrVNKJfc}}}

\end{abstract}

\begin{figure}[tb]
    \centering
    \includegraphics[width=0.97\linewidth]{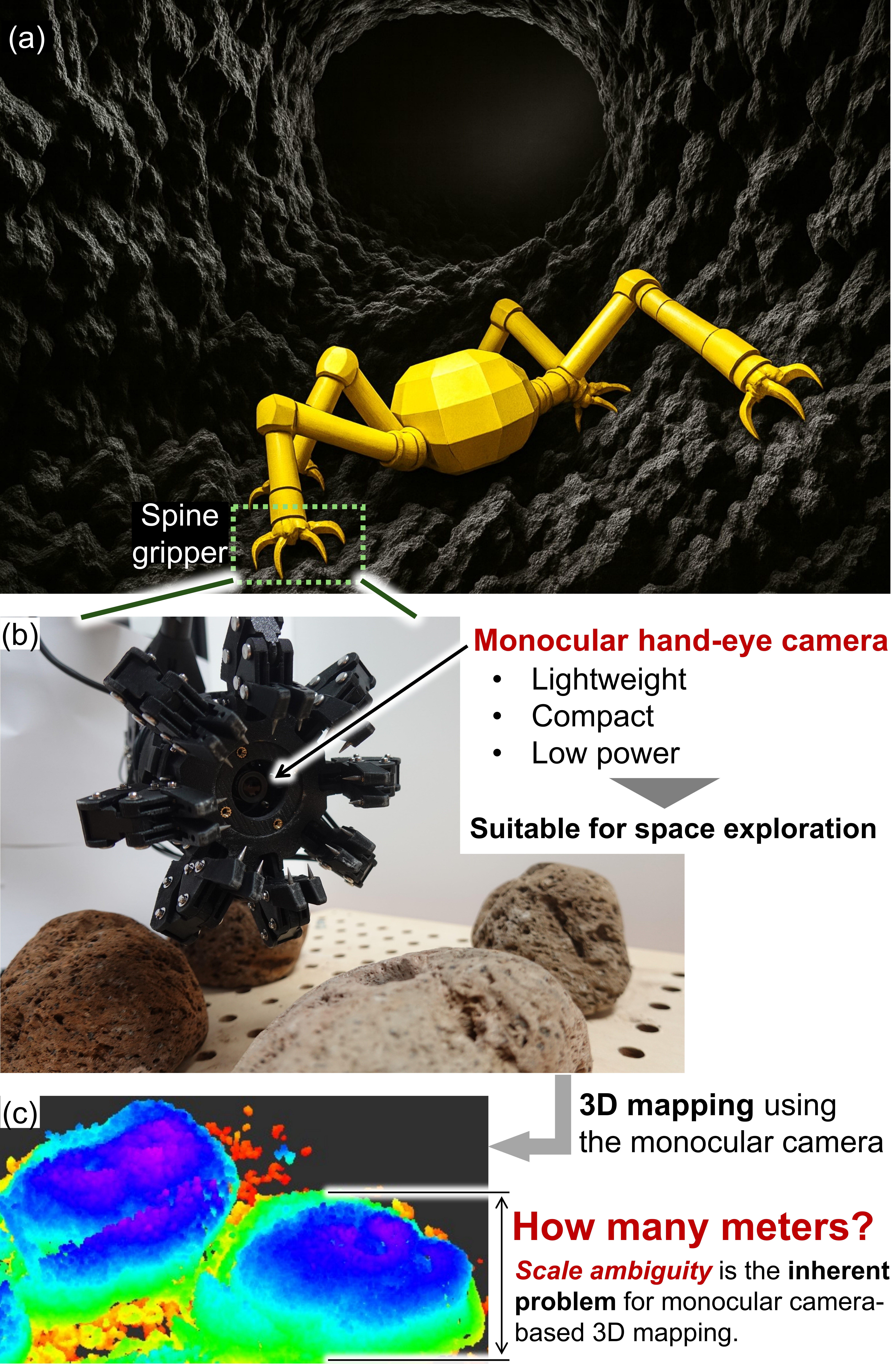}  
    \caption{(a) Limbed climbing robot exploring vertical walls such as the skylights of the Moon and Mars. (b) Although a monocular hand-eye camera enables efficient 3D mapping, (c) it inherently suffers from scale ambiguity—resulting in visually consistent 3D maps that lack metric scale information due to the limitations of monocular cameras.
             }
    \label{fig:fig1}
\end{figure}
\section{Introduction}
\subsection{Limbed climbing robots for space exploration}
Several studies have reported the existence of lava tubes on the Moon~\cite{haruyama2009possible, kaku2017detection} and Mars~\cite{cushing2015atypical}.
Exploring these environments leads to two advantages: 1) the lava tubes may contain evidence of past geological activity or life, 2) they can be used as potential habitats for humans thanks to their natural shielding from radiation and temperature fluctuations.
The entrances to the lava tubes take the form of skylights (i.e., vertical walls), as shown in Fig.~\ref{fig:fig1}(a); thus, wheeled robots cannot explore such challenging environments.

Limbed climbing robots are well-suited for exploring cliff-like walls~\cite{yoshida2002novel} because these robots use spine-type grippers mounted on their feet to grasp natural terrain surfaces, like insects.
NASA/JPL developed a large-sized limbed climbing robot whose mass is \SI{35}{kg} and each limb is 7~DOFs~\cite{parness2017lemur}; however, lightweight robots are well-suited for space exploration because their mass drastically reflects transportation costs.
In our previous work, we developed a lightweight limbed climbing robot named HubRobo~\cite{uno2021hubrobo}, equipped with a passive spine gripper~\cite{nagaoka2018passive} suitable for efficiently grasping natural rocky terrain surfaces.
Chen et al. proposed a climbing robot named ReachBot~\cite{di2024martian,chen2024locomotion}, in which each limb is extensible and equipped with a spine-type gripper at its foot tip.

An autonomous climbing system integrates various robotic technologies, including navigation (e.g., SLAM~\cite{okawara2022lunar}, gait planning~\cite{uno2019gait}, and path planning~\cite{haji2022path}) and manipulation (e.g., force control~\cite{imai2024admittance} and perception such as graspable point detection~\cite{uno2020non}).
Among them, accurate 3D position estimation of graspable points (i.e., convex terrain) is particularly crucial for reliable autonomous climbing motions because inaccurate localization of these points can result in grasp failure and potentially lead to the robot falling from the wall.
In our previous work~\cite{uno2021hubrobo}, graspable points are extracted from 3D terrain surfaces based on 1) constructing a 3D terrain map using SLAM with LiDARs or RGB-D cameras mounted on the robot body, and 2) detecting graspable points from the 3D map.
However, this method can produce non-negligible errors in the graspable point positions, potentially leading to grasp failures, because the 3D terrain map constructed by SLAM using such LiDARs or cameras inevitably includes accumulated errors.
To obtain accurate graspable point positions, a hand-eye camera is advantageous~\cite{okawara2022lunar} because it enables direct observation of the 3D position between the graspable point and the gripper, thereby avoiding the accumulation of localization errors.
Thus, both HubRobo and ReachBot are equipped with a hand-eye camera on each gripper to enhance grasping reliability.

\subsection{Monocular hand-eye camera for limbed climbing robots}
RGB-D cameras are typically used for conventional limbed climbing robots as hand-eye cameras~\cite{chen2024locomotion,okawara2022lunar} because the 3D relative position between unknown terrain surfaces and grippers is easily obtained.
However, monocular cameras (2D images) can also estimate such 3D position in real-time based on 3D reconstruction algorithms (e.g., monocular SLAM~\cite{engel2014lsd}).
We consider a monocular hand-eye camera (Fig.~\ref{fig:fig1}(b)) to be more suitable than an RGB-D camera for 3D terrain mapping, thanks to the following advantages for space exploration:
\begin{itemize}
  \item Power saving: Monocular cameras consume significantly less power than RGB-D cameras. 
  \item Lightweight and compact structure: RGB-D cameras are relatively large and, in conventional limbed climbing robot designs, have typically been mounted obliquely on the gripper due to size constraints. Embedding a compact monocular camera directly within the gripper reduces structural complexity and allows for more compact integration.
  \item  Redundancy system: Even when an RGB-D camera is employed as a hand-eye camera, the capability to accurately estimate the 3D positions of graspable points using a monocular camera enhances system redundancy, providing robustness against sensor failures or other malfunctions.
\end{itemize}
Robot systems considering the above benefits are suitable for space exploration because 1) power saving is critical for planetary exploration missions where onboard energy is limited, 2) reducing robot weight directly contributes to transportation costs, and 3) redundant systems enhance robustness and mission continuity in harsh space environments where robot repairs are difficult.

\begin{figure}[tb]
    \centering
    \includegraphics[width=1\linewidth]{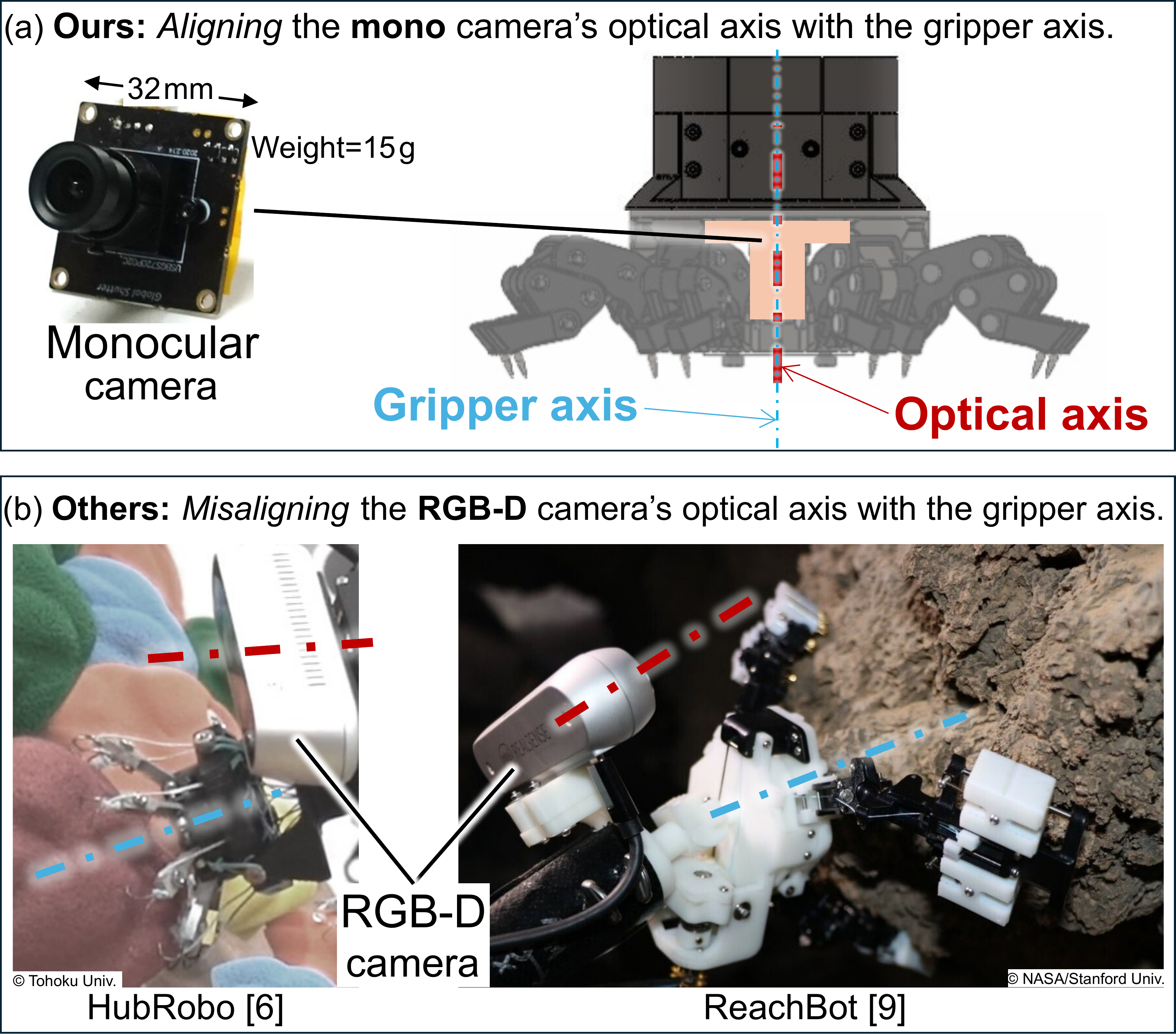}  
    \caption{Comparison between proposed and previous hand-eye camera mounting strategies. (a) In the proposed design, the monocular camera is embedded such that its optical axis aligns with the gripper axis, enabling accurate perception for grasping. (b) In previous designs (HubRobo~\cite{uno2021hubrobo}, ReachBot~\cite{chen2024locomotion}), the RGB-D camera is mounted obliquely, resulting in a misalignment between the optical and gripper axes.
             }
    \label{fig:handeye}
\end{figure}
Despite these advantages, a major limitation of monocular camera-based 3D mapping is scale ambiguity (Fig.~\ref{fig:fig1}(c)): while the 3D terrain surfaces constructed from monocular images are geometrically consistent, they inherently lack metric scale information in principle.
As a result, graspable point positions extracted from the unscaled 3D terrain surfaces cannot be directly used to control the robot’s limbs.
To estimate scaled graspable point positions based on monocular cameras, we propose a scale-aware monocular SLAM based on incorporating forward kinematics of limbed climbing robots.
The proposed method jointly estimates a scaled 3D terrain map around a gripper and the gripper's pose by fusing the limb kinematics-based and monocular camera-based gripper motion constraints.
Furthermore, our framework includes autonomous grasping based on a graspable point extracted from the scaled 3D map.

\begin{figure*}[tb]
    \centering
    \includegraphics[width=1.0\linewidth]{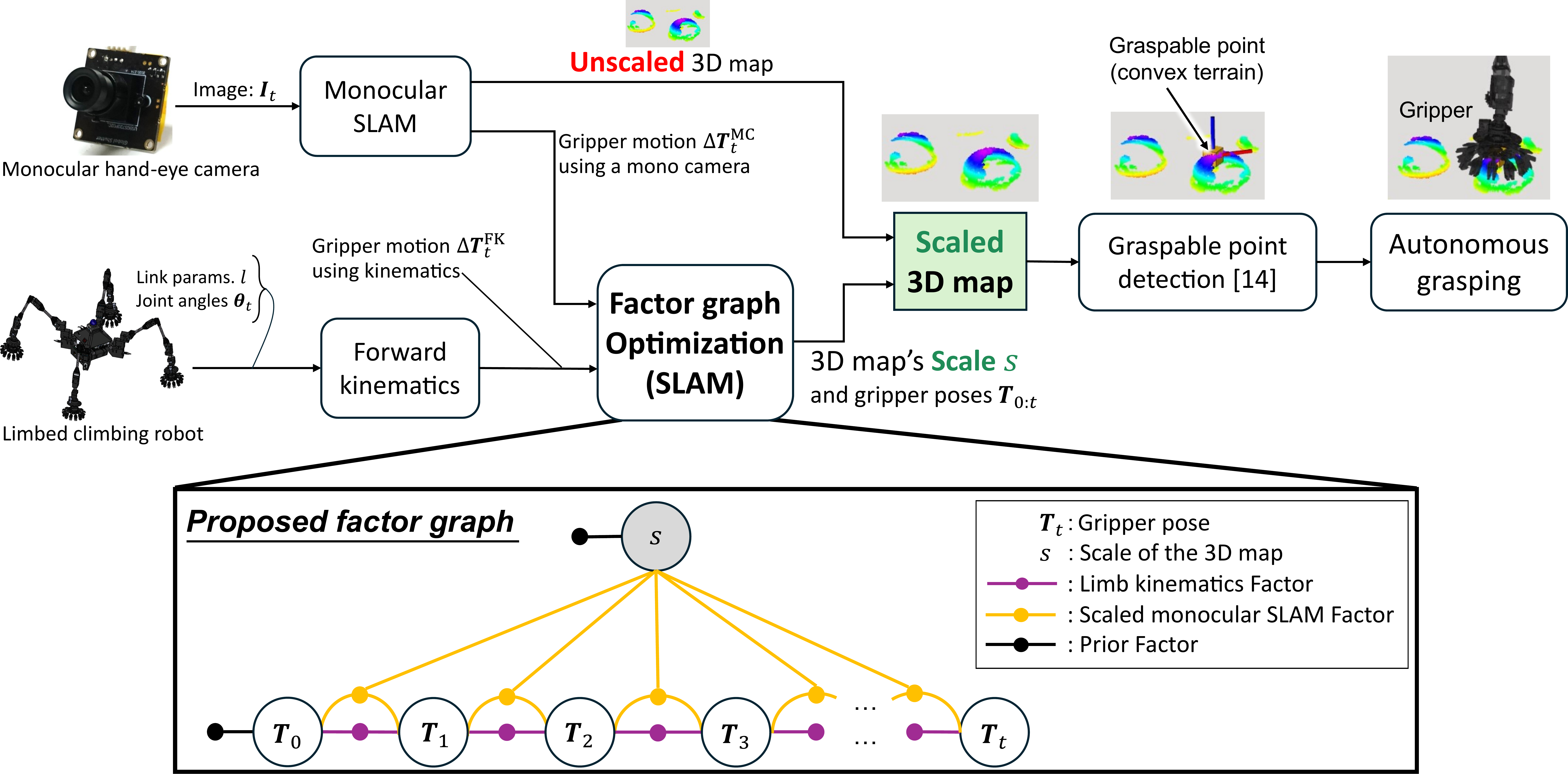}  
    \caption{Overview of the proposed framework. To construct the scaled 3D map from the unscaled 3D map, we jointly estimate the 3D map's scale $s$ and the gripper poses $\bm{T}_{0:t}$ by fusing monocular camera-based constraints (scaled monocular SLAM factor) and limb forward kinematics-based constraints (limb kinematics factor) based on factor graph optimization. Finally, we control the limb and its gripper to grasp the graspable point (convex terrain) based on the scaled 3D map.
             }
    \label{fig:overview}
\end{figure*}
In this study, our contributions are threefold:
\begin{enumerate}
  \item To deal with the scale ambiguity problem, we proposed a scale-aware monocular SLAM by fusing the limb kinematics-based and monocular camera-based motion constraints based on factor graph optimization.
  \item We developed a hand-eye system by embedding a monocular camera within the gripper and aligning its optical axis with the gripper’s vertical axis (Fig.~\ref{fig:handeye}); thus, we can simplify both the hardware design and control architecture.
  \item We demonstrated that the proposed framework enables a limbed climbing robot to automatically grasp terrain surfaces based on constructed scaled 3D maps in real-world experiments.
\end{enumerate}

\section{Related Work}
Estimating scaled target position (i.e., depth value) with a monocular camera has typically been approached in two ways: (i) learning or calibrating an appearance-to-depth converter, or (ii) pseudo-stereo vision.
Haugal{\o}kken et al.~\cite{haugalokken2020monocular} follow the former approach by using an empirically calibrated scaling function that converts the object’s image area into a depth value, for capturing the object. While simple and lightweight, this approach is object- and camera-dependent, often requiring per-target and per-camera retuning.
Horng et al.~\cite{horng2020object} take the latter way: they translate an end-effector mounting a monocular hand-eye camera by a small amount to obtain two views, compute a disparity map, and infer depth from the known motion. In their approaches, the end-effector motion is treated deterministically; the uncertainty of the \emph{baseline} (hand translation) is neither modeled nor estimated explicitly. As a result, the errors of the end-effector motion can propagate directly into a position accuracy~\cite{klingensmith2016articulated,houseago2019ko}.

We address these limitations by jointly estimating the scale and the end-effector position within a single optimization. 
Inspired by UAV visual--inertial SLAM approaches that fuse monocular vision with IMU data to recover scale~\cite{nutzi2011fusion,ullah2018emovi}, we fuse motion constraints derived from forward kinematics and monocular visual SLAM, by factor graph optimization. 
Thus, our method removes the reliance on hand-crafted scaling functions~\cite{haugalokken2020monocular} and considers robot motion errors~\cite{horng2020object}.

\section{Developed hand-eye system}
As shown in Fig.~\ref{fig:handeye}, we developed a hand-eye system that mounts a camera to a gripper.
In our design, a lightweight monocular camera (Fig.\ref{fig:handeye}(a)) is embedded inside the gripper, whereas in previous climbing robots~\cite{uno2021hubrobo}~and~\cite{chen2024locomotion}, RGB-D cameras (Fig.\ref{fig:handeye}(b)) were mounted obliquely outside the gripper.
As a hand-eye camera, we selected the ELP-USBGS720P02C-L36 (width: \SI{32}{mm}, weight: \SI{15}{g}) for its wide field of view (100~$\deg$), global shutter, and low power consumption.
Notably, its maximum power consumption is only one quarter of that of the Intel RealSense D435i used in HubRobo.
Furthermore, ReachBot is equipped with Intel RealSense D455, which is more powerful than RealSense D435i.
A hand-eye camera is mounted on each limb; thus, reducing the power consumption of each camera yields a cumulative benefit, which is particularly critical for space exploration missions.

By aligning the camera’s optical axis with the gripper’s vertical central axis, the system can observe a terrain surface to be grasped until just before grasp completion, thereby simplifying limb control for grasping.

While the developed hand-eye system is well-suited for space exploration, a monocular camera cannot inherently provide graspable point positions considering scale information due to scale ambiguity.
The following sections describe our approach to dealing with this limitation.

\section{Overview}
This section provides an overview of the proposed framework for 3D terrain mapping using a monocular hand-eye camera, aiming to enable autonomous climbing on vertical walls such as skylights of the Moon and Mars.

As illustrated in Fig.~\ref{fig:overview}, we simultaneously estimate the time-series gripper poses $\bm{T}_0, \bm{T}_1, ..., \bm{T}_t = [{\bm R}_t | {\bm t}_t] \in SE(3)$ and the scale $s$ of a 3D map built by monocular SLAM, by fusing outputs of monocular SLAM and limb forward kinematics.
This fusion addresses the scale ambiguity inherent in monocular SLAM by considering consistency between the scaled limb forward kinematics-based gripper motion $\Delta\bm{T}^{\rm{FK}}_{t}$ and the unscaled monocular camera-based motion $\Delta\bm{T}^{\rm{MC}}_{t}$.
Factor graph optimization is a smoothing approach to jointly optimize constraints defined by multiple time series measurements ($\Delta\bm{T}^{\rm{FK}}_{0:t}$, $\Delta\bm{T}^{\rm{MC}}_{0:t}$); thus, the scale $s$ is estimated by considering consistency of the multiple constraints.
The smoothing approaches contrast with typical filtering methods (e.g., Kalman filter)\cite{strasdat2012visual}, which rely solely on the most recent data.
Furthermore, combining camera-based and limb kinematics-based constraints enables more accurate 3D terrain mapping than using either modality alone~\cite{klingensmith2016articulated}. 
Note that this related work~\cite{klingensmith2016articulated} used an RGB-D camera.

Once the scale $s$ is estimated, it is applied to the unscaled 3D map produced by monocular SLAM, resulting in a metrically accurate terrain representation. 
From this scaled 3D terrain surface, graspable points (i.e., convex terrain) are identified by the graspable point detection algorithm~\cite{uno2020non}. 
The climbing robot then executes limb motions to autonomously grasp these points and climb unknown vertical walls such as the skylights.

\section{Methodology}
\subsection{Preliminaries and Problem statement}
\textbf{Factor graph optimization:} Factor graph optimization is a widely used framework for efficiently solving sensor fusion-based SLAM problems. 
A factor graph consists of variable nodes and factor nodes, where variables represent unknown states (e.g., robot poses, scale parameter), and factors represent constraints between states based on sensor measurements (e.g., 2D images, joint encoder values). 
Assuming that sensor measurement noises follow a Gaussian distribution, factor graph optimization can be formulated as a weighted nonlinear least squares problem, which minimizes an objective function defined by the sum of constraints (factors) derived from each sensor measurement.
Such optimization problems are typically solved using numerical methods such as the Gauss–Newton or Levenberg–Marquardt algorithms, both of which are well-suited for nonlinear least squares problems.

\textbf{Factor graph optimization for our SLAM formulation:} As illustrated in Fig.~\ref{fig:overview}, the scale $s$ of the 3D map and the time series gripper poses $\bm{T}_0, \bm{T}_1, ..., \bm{T}_t$ are defined as unknown states in our SLAM formulation.
Time series data of monocular images $\bm{I}_t$ and joint angles $\bm{\theta}_t$ are used to define factor nodes (constraints) as sensor observations.
To simultaneously estimate $\bm{T}_0, \bm{T}_1, ..., \bm{T}_t$ and $s$, the following objective function $e$ is minimized:
\begin{equation}
    \label{eq:FG}
    e = e^{\text{p}}(\bm{T}_{0}, s) + \sum_{i=1}^{t} e^{\text{MC}}(\bm{T}_{i-1}, \bm{T}_i, s) + \sum_{i=1}^{t} e^{\text{FK}}(\bm{T}_{i-1}, \bm{T}_i),
\end{equation}
where $e^{\text{MC}}$ and $e^{\text{FK}}$ are factors (i.e., weighted nonlinear least squares errors) defined by monocular cameras and limb forward kinematics, respectively.
In our framework, $e^{\text{MC}}$ and $e^{\text{FK}}$ are named the \textit{scaled monocular SLAM factor} and the \textit{limb kinematics factor}, respectively.
The detailed formulations of these factors are provided in the following subsections.
The prior factor $e^{\text{p}}$ (black factor node in Fig.~\ref{fig:overview}) is applied to the initial gripper pose $\bm{T}_{0}$ and the scale $s$ to fix the reference frame.

\subsection{Limb kinematics factor}
The \textit{limb kinematics factor} $e^{\text{FK}}(\bm{T}_{i-1},\bm{T}_i)$ provides motion constraints based on limb forward kinematics of climbing robots.
Forward kinematics computes scaled gripper motions from each limb's joint angles $\bm{\theta}_i$ and link parameters $\bm{l}$ (e.g., link length).
This motion constraint behaves similarly to wheel odometry-based constraints~\cite{okawara2024tightly,okawara2025ras} or leg odometry-based ones~\cite{wisth2022vilens,okawara2025ral} in typical SLAM formulations for mobile robots.
The forward kinematics-based motion $\Delta\bm{T}^{\rm{FK}}_{i}$ (relative transformation) is calculated as follows based on difference between gripper poses $\bm{T}^{\rm{FK}}_{i-1}$ and $\bm{T}^{\rm{FK}}_{i}$ at time $i-1$ and $i$, respectively:
\begin{equation}
    \label{eq:deltaFK}
    \Delta\bm{T}^{\rm{FK}}_{i} = (\bm{T}^{\rm{FK}}_{i-1})^{-1} ~\bm{T}^{\rm{FK}}_{i}.
\end{equation}

To derive the factor (i.e., weighted nonlinear least squares errors), we first define a residual $\bm{r}_{i}^{\text{FK}} \in \mathbb{R}^6$ based on difference between the state (optimization value)-based motion ($\bm{T}_{i-1}^{-1} \bm{T}_{i}$) and the observation-based motion ($\Delta\bm{T}^{\rm{FK}}_{i}$), as follows:
\begin{equation}
    \label{eq:logFK}
    \bm{r}_{i}^{\text{FK}} = \text{log} (\bm{T}_{i-1}^{-1} \bm{T}_{i} (\Delta\bm{T}^{\rm{FK}}_{i})^{-1}),
\end{equation}
where $\text{log}(\cdot)$ denotes the logarithm map that converts elements from the Lie group $SE(3)$ to its corresponding Lie algebra $se(3)$.
In other words, Eq.~(\ref{eq:logFK}) converts a homogeneous matrix-based residual (16 elements, i.e., redundant for describing a 6~DOF pose) into a 6-dimensional vector-based representation ($\bm{r}_{i}^{\text{FK}} \in \mathbb{R}^6$, i.e., non-redundant).
According to $\bm{r}_{t}^{\text{FK}}$ and its uncertainty (i.e., covariance matrix) $(\bm{\Sigma}^{\text{FK}})^{-1} \in \mathbb R^{6 \times 6}$, the limb kinematics factor $e^{\text{FK}}(\bm{T}_{i-1},\bm{T}_i)$ is finally defined as mahalanobis distance:
\begin{equation}
    \label{eq:FKfactor}
    e^{\text{FK}}(\bm{T}_{i-1},\bm{T}_i) = (\bm{r}_{i}^{\text{FK}})^\top (\bm{\Sigma}^{\text{FK}})^{-1} \bm{r}_{i}^{\text{FK}},
\end{equation}
where the covariance matrix $(\bm{\Sigma}^{\text{FK}})^{-1}$ behaves as a weight of $\bm{r}_{i}^{\text{FK}}$ during minimization of Eq.~(\ref{eq:FG}).
For simplicity, we set constant variances regarding translational and rotational elements as follows by referring to the related work~\cite{houseago2019ko}:
\begin{equation}
    (\bm{\Sigma}^{\text{FK}})^{-1} =
    \begin{bmatrix}
    (\sigma^{\text{FK}}_{\text{trans}})^{-2} \bm{I}_{3\times3} & \bm{0}_{3\times3} \\
    \bm{0}_{3\times3} & (\sigma^{\text{FK}}_{\text{rot}})^{-2} \bm{I}_{3\times3}
    \end{bmatrix}.
\end{equation}
We set $(\sigma^{\text{FK}}_{\text{trans}})^{-2}$ and $(\sigma^{\text{FK}}_{\text{rot}})^{-2}$ to $10^{-4}$ in our implementation.

\subsection{Scaled monocular SLAM factor}
Similar to the limb kinematics factor, the \textit{scaled monocular SLAM factor} $e^{\text{MC}}(\bm{T}_{i-1}, \bm{T}_i, s)$ also serves as a motion constraint; however, unlike the limb kinematics factor, this scaled monocular SLAM factor incorporates the unknown scale $s$ in addition to the gripper poses $\bm{T}_{i-1}, \bm{T}_i$.

To obtain monocular camera-based gripper motion $\Delta\bm{T}^{\rm{MC}}_{i}$ (i.e., observation of this factor), we apply a monocular SLAM algorithm (e.g., LSD-SLAM~\cite{engel2014lsd}) to monocular hand-eye camera images.
The monocular SLAM algorithm outputs time series gripper poses such as $\bm{T}^{\rm{MC}}_{i-1}$ and $\bm{T}^{\rm{MC}}_{i}$; thus, we calculate the monocular hand-eye camera-based gripper motion $\Delta\bm{T}^{\rm{MC}}_{i}$ similar to Eq.~(\ref{eq:deltaFK}) as follows:
\begin{equation}
    \label{eq:deltaMONO}
    \Delta\bm{T}^{\rm{MC}}_{i} = (\bm{T}^{\rm{MC}}_{i-1})^{-1} ~\bm{T}^{\rm{MC}}_{i}.
\end{equation}
Note that $\Delta\bm{T}^{\rm{MC}}_{i}$ is not scaled in translational elements due to the inherent scale ambiguity of monocular vision, in contrast to Eq.~(\ref{eq:deltaFK}).
To estimate $s$ in translational elements by $\Delta\bm{T}^{\rm{MC}}_{i}$, we decompose $\bm{T} \in SE(3)$ into translational elements $\bm{t}\in\mathbb{R}^3$ and rotational elements $\bm{R} \in {SO(3)}$.

First, we derive the rotational residual $\bm{r}_{\text{rot},i}^{\text{MC}} \in so(3)$ because the rotational elements are independent of the scale $s$.
Since the logarithm map can be applied to $\bm{R}$ as well as to $\bm{T}$, the rotational residual is defined as follows:
\begin{equation}
    \label{eq:logR}
    \bm{r}_{\text{rot},i}^{\text{MC}} = \text{log} (\bm{R}_{i-1}^{-1} \bm{R}_{i} (\Delta\bm{R}^{\rm{MC}}_{i})^{-1}).
\end{equation}
Next, the translational residual $\bm{r}_{\text{trans},i}^{\text{MC}}$ is defined by scaling the monocular SLAM-based translational motion $\Delta\bm{t}^{\rm{MC}}_{i}$ using the scale parameter $s$:
\begin{equation}
    \label{eq:r_trans}
    \bm{r}_{\text{trans},i}^{\text{MC}} = (\bm{t}_{i} - \bm{t}_{i-1}) - s \Delta\bm{t}^{\rm{MC}}_{i}.
\end{equation}
Finally, the scaled monocular SLAM factor $e^{\text{MC}}(\bm{T}_{i-1}, \bm{T}_i, s)$ is also defined as mahalanobis distance:
\begin{align}
    \label{eq:MONOFACTOR}
    e^{\text{MC}}(\bm{T}_{i-1},\bm{T}_i, s) 
        &= (\bm{r}_{i}^{\text{MC}})^\top (\bm{\Sigma}^{\text{MC}})^{-1} \bm{r}_{i}^{\text{MC}}, \\
    \bm{r}_{i}^{\text{MC}} 
        &= 
        \begin{bmatrix}
            \bm{r}_{\text{trans},i}^{\text{MC}} \\
            \bm{r}_{\text{rot},i}^{\text{MC}}
        \end{bmatrix},
\end{align}
where $(\bm{\Sigma}^{\text{MC}})^{-1}$ is uncertainty (i.e., covariance matrix) of $\bm{r}_{i}^{\text{MC}}$ to serve as a weighting matrix in factor graph optimization:
\begin{equation}
    (\bm{\Sigma}^{\text{MC}})^{-1} =
    \begin{bmatrix}
    (\sigma^{\text{MC}}_{\text{trans}})^{-2} \bm{I}_{3\times3} & \bm{0}_{3\times3} \\
    \bm{0}_{3\times3} & (\sigma^{\text{MC}}_{\text{rot}})^{-2} \bm{I}_{3\times3}
    \end{bmatrix}.
\end{equation}
We set $(\sigma^{\text{MC}}_{\text{trans}})^{-2}$ and $(\sigma^{\text{MC}}_{\text{rot}})^{-2}$ to $10^{-3}$.

\subsection{Implementation details}
The proposed factor graph (i.e., Fig.\ref{fig:overview}, Eq.(\ref{eq:FG})) is constructed by incrementally accumulating factors (Eq.(\ref{eq:FKfactor}), Eq.(\ref{eq:MONOFACTOR})) at each time step. 
Once a sufficient number of factors (i.e., multi-perspective images of terrain surfaces to be grasped have been obtained) are accumulated, Eq.~(\ref{eq:FG}) is optimized using the Levenberg–Marquardt method to jointly estimate the scale parameter $s$ and the time-series gripper poses. 
The proposed factor graph and its optimization are implemented using GTSAM\footnote{https://github.com/borglab/gtsam}, which is a de facto standard library for factor graph optimization related to SLAM.

We used Large-Scale Direct (LSD) SLAM~\cite{engel2014lsd} as a monocular SLAM approach to obtain unscaled 3D terrain maps and gripper poses.
We consider LSD-SLAM to be suitable for space exploration scenarios because this algorithm can efficiently construct a dense 3D map using only a CPU, in contrast to GPU-dependent methods~\cite{whelan2015elasticfusion} (i.e., high power consumption) or feature-based methods~\cite{mur2015orb} (i.e., building a sparse 3D map, which is difficult to extract graspable points from this type of map).

To identify graspable points from a 3D terrain map, we used a graspable point detection algorithm~\cite{uno2020non}, which is suited for spine-type grippers. First, we construct a terrain array by voxelizing the map and labeling each voxel as 0 (free) or 1 (occupied). Next, we define an approximately bowl-shaped 3D binary mask determined by the gripper’s geometry (e.g., finger range of motion and overall size). Finally, we slide the gripper mask over the terrain array; a location is regarded as a graspable point if all solid voxels of the mask are supported by occupied terrain voxels. More details of this algorithm are shown in~\cite{uno2020non}.

\begin{figure}[tb]
    \centering
    \includegraphics[width=1\linewidth]{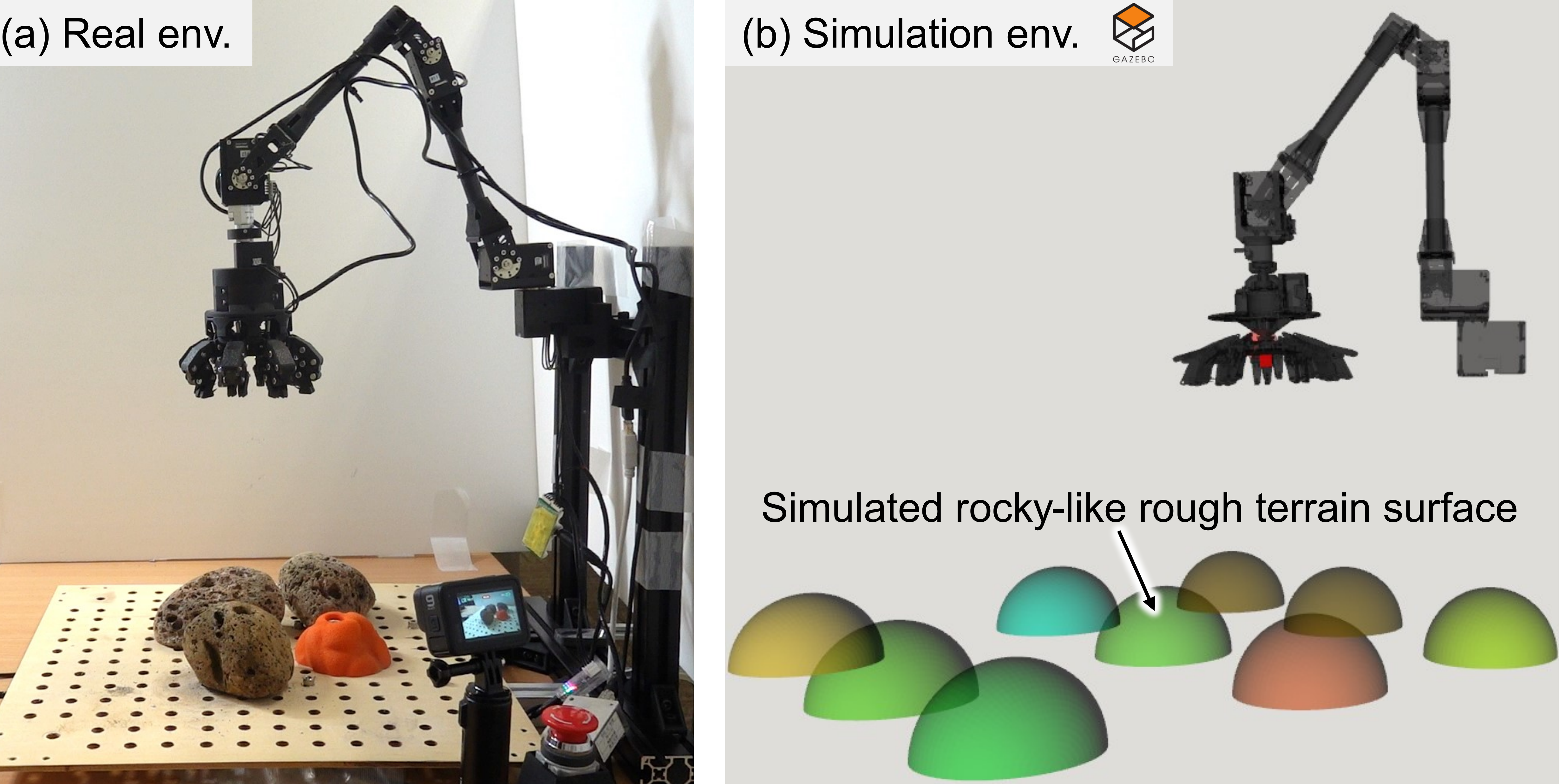}  
    \caption{Testbed used for (a) real-world experiment and (b) simulation experiment.
             }
    \label{fig:testbed}
\end{figure}

\begin{figure}[tb]
    \centering
    \includegraphics[width=1\linewidth]{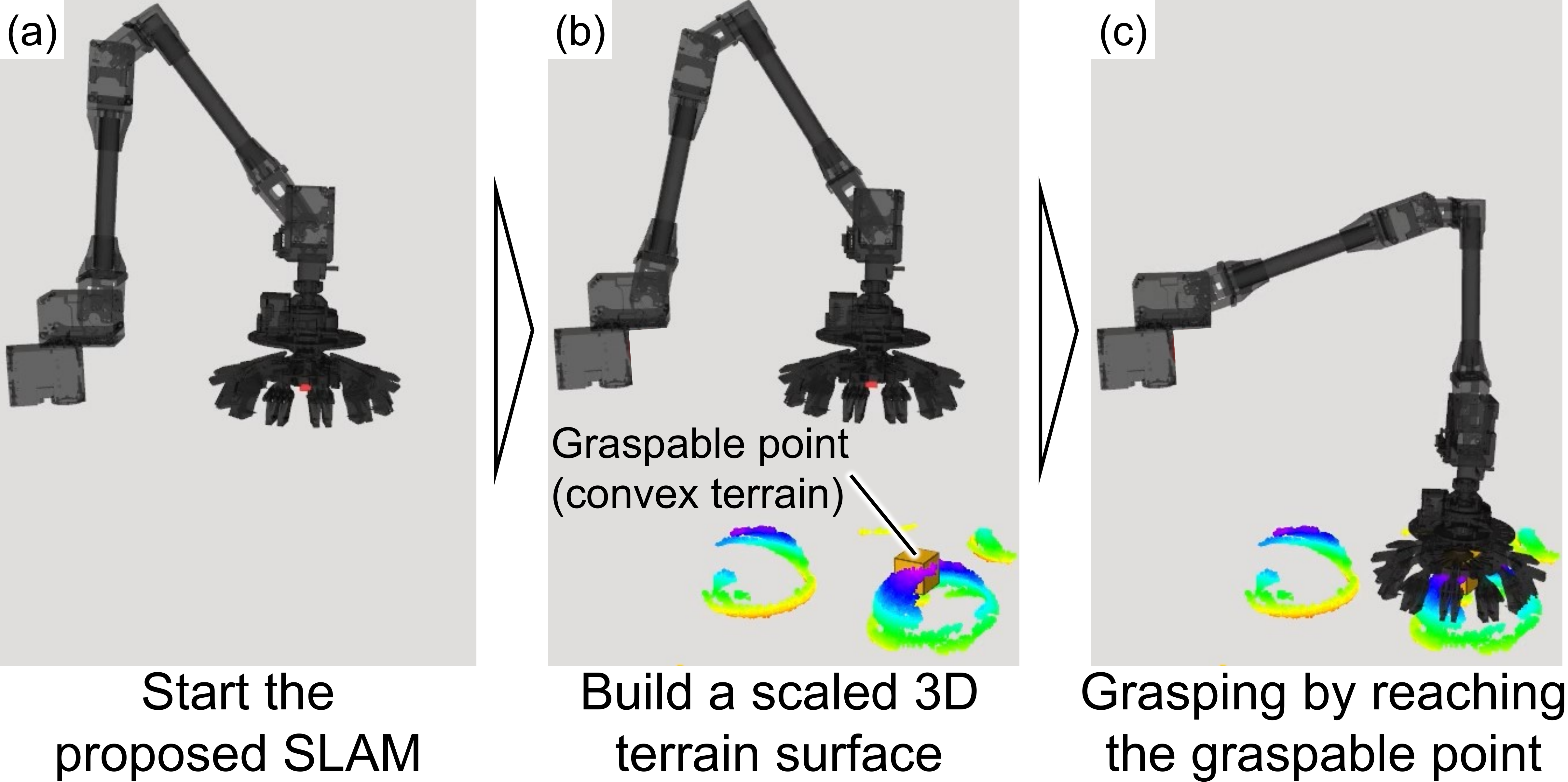}  
    \caption{Snapshots of our framework in the simulation experiment. The scaled 3D terrain surface was constructed by our SLAM method, and the gripper reached the graspable point extracted from the 3D map.
             }
    \label{fig:sim_result}
\end{figure}
\section{Experimental results}
\subsection{Experimental setup}
We evaluated the proposed framework (Fig.~\ref{fig:overview}) through both physics-based simulations and real-world experiments. 
As shown in Fig.~\ref{fig:testbed}, a single limb of our climbing robot was used as a testbed to demonstrate the proof of concept of our framework. 
Our testbed has a 4 DOF limb and is equipped with a spine-type gripper designed to grasp natural rocky terrain surfaces. 
A lightweight monocular camera was embedded within the gripper (Fig.~\ref{fig:handeye}(a)), with its optical axis aligned along the vertical center axis of the gripper. 
The hand-eye camera streams monocular images at 30 Hz, which are processed by a monocular SLAM (e.g., LSD-SLAM).
We implemented our framework using ROS 2 Humble on Ubuntu 22.04.

\subsection{Experimental validation in a simulation environment}
We demonstrated that the gripper automatically reaches a graspable point (i.e., convex terrain) based on our SLAM algorithm in the Gazebo simulator.
To simulate rough terrain surfaces similar to rocky terrain, we placed multiple hemispheres on the ground, as shown in Fig.~\ref{fig:testbed}(b).
Following our framework (Fig.~\ref{fig:overview}), we first executed the proposed SLAM algorithm (Fig.~\ref{fig:sim_result}(a)) by minimizing Eq.~(\ref{eq:FG}), thereby constructing a scaled 3D terrain surface (Fig.\ref{fig:sim_result}(b)).
From this reconstructed 3D map, graspable points were identified using the graspable point detection algorithm\cite{uno2020non}, as shown in Fig.\ref{fig:sim_result}(b).
Finally, we demonstrated that the gripper automatically reached the graspable point, as shown in Fig.~\ref{fig:sim_result}(c).

To assess the accuracy of the proposed SLAM, we measured the positional error between the gripper’s palm center and the actual terrain surface after the gripper reached the graspable point.
The average positional error was $1.8 \pm 0.7$\si{mm} in six trials.
Note that force feedback control (i.e., position correction using force sensors) is not incorporated into our framework in this work because this paper focuses on hand-eye camera-based 3D mapping.
These results indicate that the proposed SLAM successfully reconstructed a scaled 3D terrain surface using a monocular hand-eye camera by fusing monocular camera-based and limb kinematics-based constraints.

\subsection{Experimental validation in a real-world environment}
We conducted real-world experiments (Fig.~\ref{fig:testbed}(a)) to demonstrate that the gripper can reliably grasp convex terrain surfaces through our proposed framework.
Grasping with multiple contacts of spines is a complex phenomenon; thus, we cannot observe whether the gripper successfully grasps the convex terrain surfaces in the simulation environments.
Therefore, in the real experiment, grasp success was determined by comparing the measured gripping force against a predefined threshold.
The gripping force was measured using a force sensor (Leptrino CFS018CA201U) mounted on the gripper.

\begin{figure}[tb]
    \centering
    \includegraphics[width=1\linewidth]{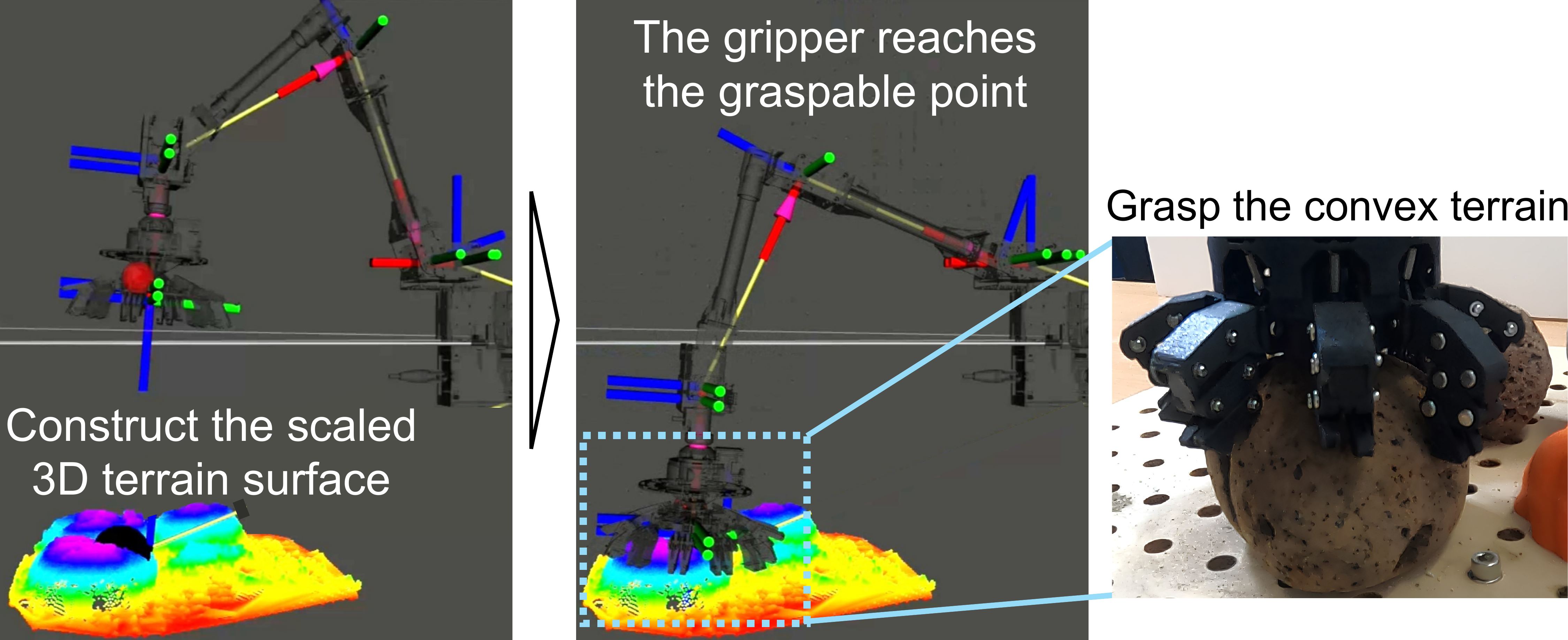}  
    \caption{Snapshots of our framework in the real experiment. The gripper grasped the convex terrain surface based on the scaled 3D terrain map constructed by our SLAM method. Note that these snapshots are visualized from real robot data (i.e., these snapshots are not obtained from simulators).
             }
    \label{fig:real_result}
\end{figure}

\begin{figure}[tb]
    \centering
    \includegraphics[width=1\linewidth]{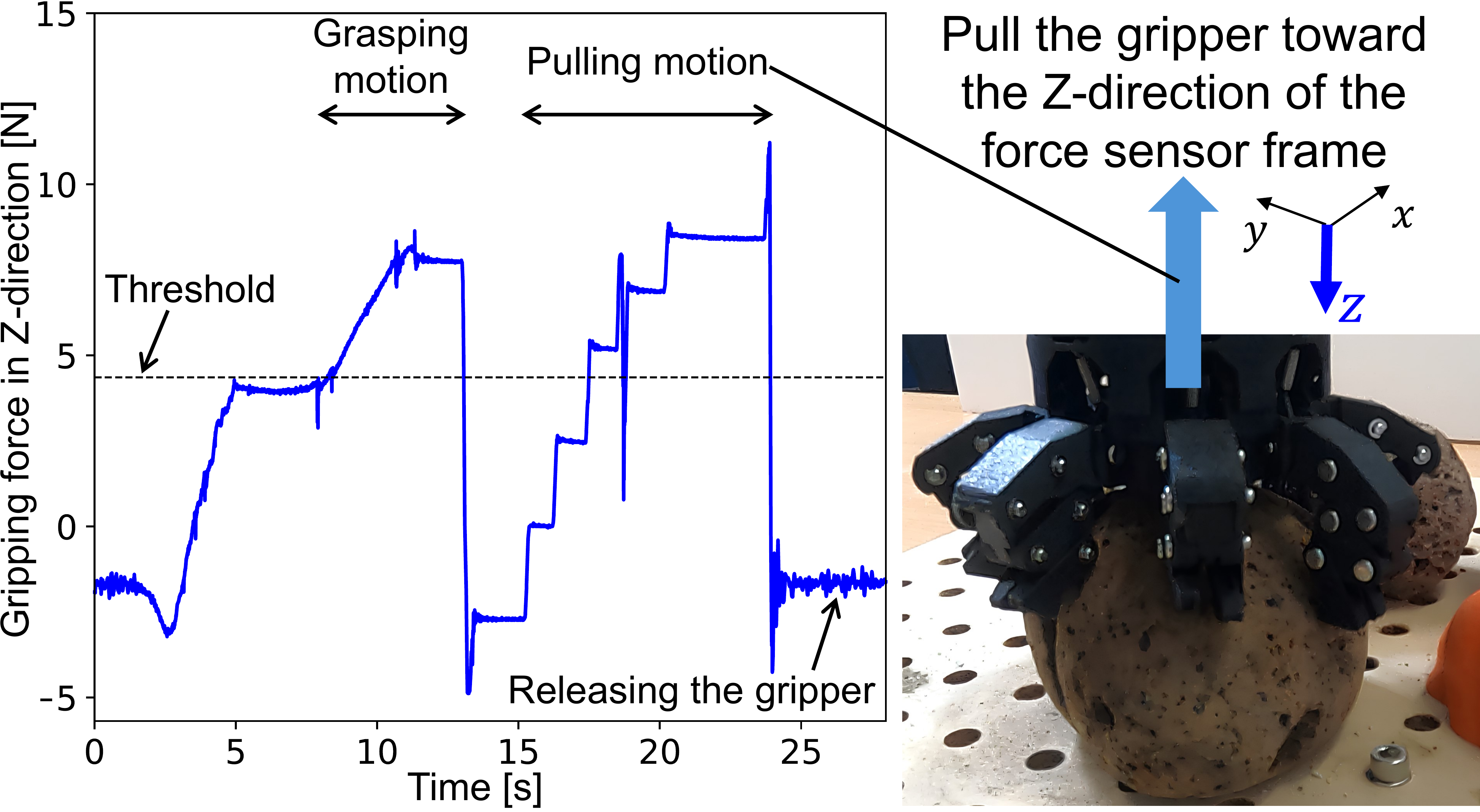}  
    \caption{Time series data of the Z-direction's force sensor values (i.e., gripping force) during our framework.
             The force exceeded the threshold; thus, the gripping force was sufficient to grasp the graspable point, and this autonomous grasping (Fig.~\ref{fig:real_result}) was successful.
             }
    \label{fig:fs}
\end{figure}

We show the snapshots of this experiment in Fig.~\ref{fig:real_result}.
Similarly to the simulation experiment, we can see that our SLAM algorithm successfully constructed the scaled 3D terrain surface and the gripper automatically reached the graspable point even in the real-world experiment.
In contrast to the simulation experiment, the gripper was pulled to determine whether the gripper properly grasps the convex terrain surface after completing the grasping motion.
Fig.~\ref{fig:fs} presents the time-series data of the gripping force in the $Z$-direction during this experiment.
We can see that the gripping force exceeded the threshold; thus, the gripping force is sufficient to grasp the convex terrain surface with our spine gripper.
The threshold was set under two assumptions: 1) four single-limb units are mounted on a robot body similar to HubRobo and ReachBot, and 2) the sum of the gripping forces must be sufficient to support the total robot mass under the lunar gravity condition.
Therefore, we demonstrated that our SLAM algorithm constructed the scaled 3D terrain surface using the monocular hand-eye camera, and the gripper properly grasped the convex terrain surface based on our framework, even in real-world conditions.


\section{Conclusion}
This work presented a 3D terrain mapping system for limbed climbing robots, leveraging lightweight monocular hand-eye cameras embedded within the grippers.
By fusing monocular camera-based and limb kinematics-based constraints within the proposed factor graph, our SLAM algorithm deals with the inherent scale ambiguity of monocular vision, thereby constructing scaled 3D terrain surfaces (i.e., 3D map). 
In simulation experiments, we demonstrated that the gripper reached the graspable point (convex terrain) with high accuracy (positional error was $1.8 \pm 0.7$~\si{mm}) based on our framework.
In real-world experiments, we verified that the gripper can automatically grasp the graspable point by confirming that the gripping force exceeded the threshold.
Therefore, these results demonstrate that compact, lightweight, and low-power monocular cameras can serve as effective hand-eye cameras for limbed climbing robots in space exploration scenarios.
Our framework is broadly applicable to hand-eye systems of various limbed climbing robots because monocular image acquisition and forward kinematics computation are fundamental capabilities of hand-eye systems.

In this study, the framework was validated using a single-limb robot as a proof of concept.
Future work will focus on applying the framework to a four-limbed climbing robot and evaluating its performance in more complex and realistic experimental environments~\cite{uno2024lower}.
Furthermore, we plan to integrate the proposed method with other sensors (e.g., LiDAR, IMU) mounted on the robot base.
In particular, we will extend the factor graph to jointly optimize the motion constraints associated with each gripper and the base in a consistent manner.


\balance

\bibliographystyle{IEEEtran}
\bibliography{isparo2025_bib}

\end{document}